\documentclass[runningheads]{llncs}

\usepackage{graphicx}
\usepackage{amsmath}
\usepackage{amssymb}
\usepackage{booktabs}
\usepackage{multirow}
\usepackage[table]{xcolor}
\usepackage{orcidlink}

\definecolor{hilnngray}{RGB}{238,238,238}
\definecolor{bestgreen}{RGB}{0,100,0}
\definecolor{citewine}{RGB}{145,32,48}
\definecolor{linkblue}{RGB}{25,70,130}

\newcommand{\best}[1]{\textbf{\textcolor{bestgreen}{#1}}}
\newcommand{\ourscell}[1]{\cellcolor{hilnngray}#1}

\hypersetup{
    colorlinks=true,
    citecolor=citewine,
    linkcolor=linkblue,
    urlcolor=citewine
}

\begin{document}

\title{History-informed Lagrangian Neural Networks}

\author{
Tianshuo Zhang\orcidlink{0000-0002-7920-4473} \and
Xianglei Xing\orcidlink{0000-0002-4159-1922}
\thanks{Corresponding author: xingxl@hrbeu.edu.cn} \and
Wenzhe Zhai\orcidlink{0000-0003-0996-6832} \and
Jia Gao\orcidlink{0009-0007-7107-7610} \and
He Cao\orcidlink{0009-0000-2282-966X}
}

\authorrunning{T. Zhang et al.}

\institute{
College of Intelligent Systems Science and Engineering,\\
Harbin Engineering University, Harbin 150001, China
}

\maketitle             
\begin{abstract}
Forecasting the long-horizon evolution of mechanical systems from position-only observations is a pivotal yet difficult task, as hidden velocities and trajectory-specific physical properties must be inferred simultaneously. Although physics-guided neural networks like Lagrangian Neural Networks (LNNs) guarantee physical plausibility, they generally require complete state inputs and lack adaptability to changing system parameters. To break these limitations, we introduce History-informed Lagrangian Neural Networks (HiLNN). Grounded in the insight that temporal position sequences implicitly encode underlying dynamics, HiLNN employs a recurrent encoder to extract a latent context from history. This context not only reconstructs the unobserved initial velocity but also adaptively modulates the mass matrix, potential energy, and damping coefficients of a structured Lagrangian system. By leveraging a differentiable RK4 rollout scheme, the entire pipeline is optimized end-to-end under multi-step trajectory supervision and energy-consistency regularization. Empirical evaluations across conservative, dissipative, and heterogeneous variable-parameter systems show that HiLNN delivers superior long-term prediction accuracy and maintains precise energy profiles compared to state-of-the-art baselines. The source code is publicly available at \url{https://github.com/yingtian22/History-informed-LNN}.
\end{abstract}

\begin{figure}[t]
    \centering
    \includegraphics[width=\linewidth]{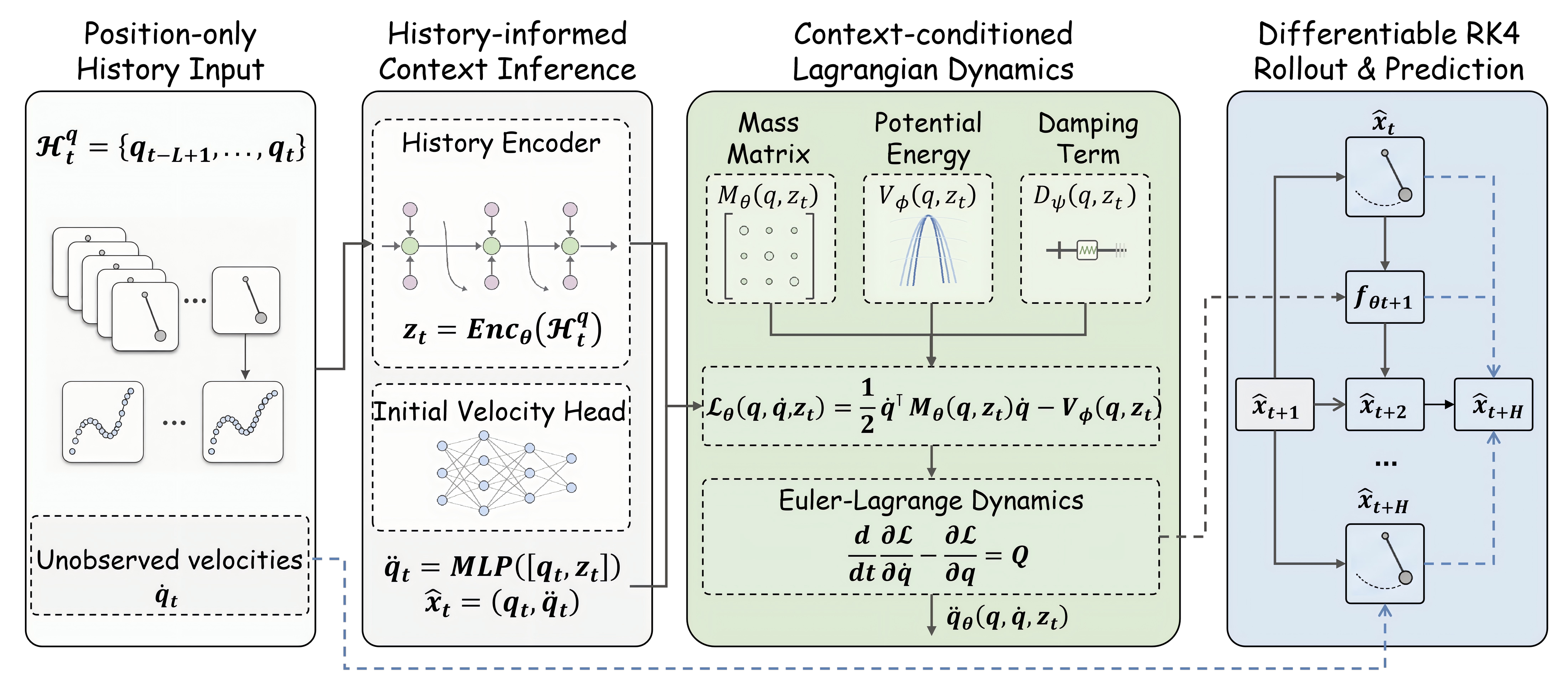}
    \caption{
    Overview of HiLNN.
    }
    \label{fig:method_overview}
\end{figure}
\section{Introduction}

Long-horizon forecasting of mechanical systems is fundamental to physical reasoning, robotics, control, and scientific modeling~\cite{battaglia2016interaction,nagabandi2018neural,sanchez2020learning,li2025trajectory,li2026evolving}.
A reliable model should remain stable and physically plausible during extended open-loop rollouts, where small errors can recursively accumulate and produce unstable trajectories or inconsistent physical quantities such as velocity and energy.

Recent neural dynamics models learn physical evolution from data through discrete-time transitions or continuous-time vector fields~\cite{nagabandi2018neural,chen2018neural,rubanova2019latent,li2025frequency,li2026distilling}.
However, they often treat dynamics as black-box mappings without explicit mechanical structure, making long-term rollouts prone to error accumulation and physical inconsistency~\cite{sanchez2020learning,greydanus2019hamiltonian}.

Physics-guided neural dynamics address this limitation by incorporating mechanical priors.
Hamiltonian and Lagrangian Neural Networks derive dynamics from learned energy functions or Lagrangians, improving interpretability and rollout stability~\cite{greydanus2019hamiltonian,cranmer2020lagrangian,lutter2019deep,finzi2020simplifying,zhong2020symplectic}.
Recent extensions further handle constrained, controlled, or dissipative systems through explicit structure and energy-dissipation mechanisms~\cite{lutter2019deep,zhong2020symplectic,sosanya2022dissipative,desai2021port}.
However, most Lagrangian models require full initial states, especially velocity, and learn a fixed global dynamics function, limiting their use under position-only observations and trajectory-varying dynamics.

Our key observation is that position history contains cues about missing states and trajectory-specific properties, consistent with delay-coordinate reconstruction and recent studies on partial-observation dynamics learning~\cite{takens1981detecting,buissonfenet2023recognition}.
Although velocity and physical parameters are unobserved, temporal position evolution reflects local motion trends and hidden dynamical variations.
We therefore infer a compact latent context from history, enabling structured dynamics to adapt to each trajectory beyond current-position or finite-difference estimates.

To this end, we propose HiLNN, a history-informed Lagrangian framework for position-only mechanical forecasting.
Given a coordinate history, HiLNN uses a recurrent encoder to infer a latent dynamical context, which estimates the missing initial velocity and conditions context-dependent mass, potential, and optional damping terms.
Starting from the inferred state, HiLNN performs differentiable RK4 rollout and jointly trains the encoder and structured dynamics module.

We evaluate HiLNN on conservative, dissipative, and variable-parameter pendulum systems.
Results show that HiLNN improves rollout accuracy and physical consistency over representative baselines, including LNN, HNN, Neural ODE, and MLP predictors.
Our contributions are summarized as follows:
\begin{itemize}
    \item We study position-only long-horizon mechanical forecasting, where hidden velocity and trajectory-specific dynamics are inferred from history.
    \item We propose HiLNN, a history-informed Lagrangian framework that conditions mass, potential, and optional damping on latent context.
    \item We show that HiLNN produces more accurate and physically consistent rollouts on conservative, dissipative, and variable-parameter systems.
\end{itemize}
\section{Related Work}

\subsection{Neural Dynamics Learning}

Neural networks, such as recurrent models and Neural ODEs~\cite{chen2018neural,rubanova2019latent,li2025ddtime}, are widely used to model continuous or discrete dynamical systems. Despite their flexibility in predicting future evolution via numerical integration, these black-box models lack explicit mechanical structure, energy consistency, or physical constraints. Consequently, they often suffer from accumulated rollout errors and physically implausible predictions over long horizons. To address this, our work builds on structured physical dynamics and introduces history-dependent context to improve long-horizon forecasting under incomplete observations.

\subsection{Physics-Guided Neural Dynamics}

Physics-guided neural dynamics embed mechanical structures to improve prediction accuracy. Representative examples include Hamiltonian Neural Networks for energy conservation~\cite{greydanus2019hamiltonian} and Lagrangian Neural Networks using the Euler--Lagrange equation~\cite{cranmer2020lagrangian}. Further extensions introduce mass matrices, dissipation, or domain priors to enhance interpretability~\cite{lutter2019deep,zhang2025floating}. However, these methods typically learn a single global model and require complete state observations. In contrast, our work conditions structured Lagrangian dynamics on a history-inferred latent context, adapting to partially observed and trajectory-dependent systems.

\subsection{Dynamics Learning from Partial Observations}

Physical dynamics are frequently observed through incomplete measurements, omitting key variables like velocity or damping. Existing approaches employ recurrent encoders, latent state-space models, or neural ODE variants to infer hidden states from observation histories~\cite{krishnan2015deep,rubanova2019latent,yildiz2019ode2vae}. Although effective, their latent dynamics remain black-box and lack explicit mechanical structure. Conversely, our method infers a latent mechanical context from position-only history to condition a Lagrangian dynamics model, enabling structured, long-horizon prediction under partial observability and trajectory-dependent variations.
\section{Method}

\subsection{Problem Formulation}
\label{sec:problem_formulation}

We study long-horizon forecasting of mechanical systems from position-only observations.
Let \(q_t \in \mathbb{R}^{d}\) be the generalized coordinate and \(\dot q_t \in \mathbb{R}^{d}\) be the corresponding velocity, forming the physical state \(x_t=(q_t,\dot q_t)\).
Unlike standard Lagrangian Neural Networks that usually assume access to the full initial state, we consider a partially observed setting where only a history of positions is available:
\begin{equation}
    \mathcal{H}_t^q = \{q_{t-L+1}, q_{t-L+2}, \ldots, q_t\}.
\end{equation}
Given \(\mathcal{H}_t^q\), the goal is to predict the future state trajectory
\begin{equation}
    \hat{\mathcal{X}}_{t+1:t+H}
    =
    \{(\hat q_{t+1}, \hat{\dot q}_{t+1}), \ldots,
    (\hat q_{t+H}, \hat{\dot q}_{t+H})\}
\end{equation}
over an \(H\)-step open-loop rollout.

This setting requires inferring both the missing initial velocity and trajectory-specific dynamics from position history.
HiLNN addresses this by using a history-informed context to condition a structured Lagrangian dynamics model.

\subsection{Overview of HiLNN}
\label{sec:overview}

HiLNN combines history-based state inference and structured Lagrangian dynamics.
As shown in Fig.~\ref{fig:method_overview}, the position history \(\mathcal{H}_t^q\) is encoded into a latent context
\begin{equation}
    z_t = \mathrm{Enc}_\theta(\mathcal{H}_t^q),
\end{equation}
which summarizes trajectory-specific dynamical information.
The context is used by a velocity head to infer the missing initial velocity \(\hat{\dot q}_t\), forming
\begin{equation}
    \hat{x}_t = (q_t, \hat{\dot q}_t).
\end{equation}
It also conditions the Lagrangian module through context-dependent mass, potential, and optional damping terms.
Starting from \(\hat{x}_t\), HiLNN performs an open-loop RK4 rollout for \(H\) steps, with \(z_t\) kept fixed as the inferred context of the current trajectory.

Unlike conventional LNNs that require known velocities and use a fixed global dynamics model, HiLNN infers both hidden velocity and trajectory-specific context from history.

\subsection{History-Informed Context Encoder}
\label{sec:history_encoder}

The history encoder maps the position history to a compact latent context.
Given
\begin{equation}
    \mathcal{H}_t^q = \{q_{t-L+1}, q_{t-L+2}, \ldots, q_t\},
\end{equation}
we encode the sequence with a recurrent network:
\begin{equation}
    h_t = \mathrm{GRU}_\theta(\mathcal{H}_t^q),
\end{equation}
where \(h_t\) is the final hidden representation.
It is projected to the latent context:
\begin{equation}
    z_t = \mathrm{MLP}_z(h_t).
\end{equation}

The context \(z_t\) captures information unavailable from the current position alone, including motion trends, hidden velocity cues, and parameter variations.
It is shared by the velocity head and the Lagrangian module, and remains fixed during rollout.

\subsection{Context-Conditioned Lagrangian Dynamics}
\label{sec:context_lagrangian}

To preserve mechanical structure, HiLNN models dynamics with a context-conditioned Lagrangian.
Given \(x=(q,\dot q)\) and context \(z_t\), we define
\begin{equation}
    \mathcal{L}_\theta(q,\dot q,z_t)
    =
    T_\theta(q,\dot q,z_t) - V_\phi(q,z_t),
\end{equation}
where \(T_\theta\) and \(V_\phi\) denote the kinetic and potential energy terms, respectively.
The kinetic energy is parameterized by a positive mass matrix:
\begin{equation}
    T_\theta(q,\dot q,z_t)
    =
    \frac{1}{2}\dot q^\top M_\theta(q,z_t)\dot q,
\end{equation}
and the potential energy is predicted by a neural network conditioned on both the coordinate and the inferred context:
\begin{equation}
    V_\phi(q,z_t) = \mathrm{MLP}_V([q,z_t]).
\end{equation}
The mass term is constrained positive.
For one-dimensional systems, we use
\begin{equation}
    M_\theta(q,z_t)
    =
    \mathrm{softplus}\big(\mathrm{MLP}_M([q,z_t])\big) + \epsilon,
\end{equation}
where \(\epsilon\) ensures numerical stability.
For multi-dimensional systems, we use a positive diagonal mass matrix with element-wise softplus.

The resulting acceleration is derived from the Euler--Lagrange equation:
\begin{equation}
    \frac{d}{dt}
    \frac{\partial \mathcal{L}_\theta}{\partial \dot q}
    -
    \frac{\partial \mathcal{L}_\theta}{\partial q}
    =
    Q,
\end{equation}
where \(Q\) is the generalized non-conservative force.
We set \(Q=0\) for conservative systems and optionally use context-conditioned damping for dissipative systems:
\begin{equation}
    Q = -D_\psi(q,z_t)\dot q,
\end{equation}
where \(D_\psi(q,z_t)\) is constrained to be non-negative.

In practice, the acceleration can be obtained by solving the Euler--Lagrange system:
\begin{equation}
    \ddot q
    =
    A^{-1}
    \left(
    \frac{\partial \mathcal{L}_\theta}{\partial q}
    -
    B\dot q
    +
    Q
    \right),
\end{equation}
where
\begin{equation}
    A =
    \frac{\partial^2 \mathcal{L}_\theta}{\partial \dot q^2},
    \qquad
    B =
    \frac{\partial^2 \mathcal{L}_\theta}{\partial q \partial \dot q}.
\end{equation}
Thus, dynamics are computed from a structured energy-based formulation rather than direct acceleration regression.
Conditioning mass, potential, and damping on \(z_t\) allows HiLNN to adapt to trajectory-dependent dynamics while retaining mechanical structure.

\subsection{Initial State Inference and Differentiable RK4 Rollout}
\label{sec:initial_state_rollout}

Since the initial velocity is unobserved, HiLNN infers it from the last position and context:
\begin{equation}
    \hat{\dot q}_t
    =
    \mathrm{MLP}_{v}([q_t,z_t]),
\end{equation}
where \([q_t,z_t]\) denotes concatenation.
The initial state is then
\begin{equation}
    \hat{x}_t = (q_t,\hat{\dot q}_t).
\end{equation}

Starting from \(\hat{x}_t\), the model performs open-loop prediction using the context-conditioned dynamics.
Let
\begin{equation}
    \frac{d x}{dt}
    =
    f_\theta(x,z_t)
    =
    \begin{bmatrix}
        \dot q \\
        \ddot q_\theta(q,\dot q,z_t)
    \end{bmatrix},
\end{equation}
where \(\ddot q_\theta\) is derived from the context-conditioned Euler--Lagrange dynamics.
We use a fourth-order Runge--Kutta integrator to advance the state:
\begin{equation}
    \hat{x}_{t+k+1}
    =
    \mathrm{RK4}\big(f_\theta,\hat{x}_{t+k},z_t,\Delta t\big),
    \quad k=0,\ldots,H-1.
\end{equation}

The rollout is fully differentiable, with \(z_t\) fixed while the state is recursively updated by the learned dynamics.

\subsection{Training Objective}
\label{sec:training_objective}

HiLNN is trained end-to-end with multi-step rollout supervision.
Given predicted trajectory \(\hat{\mathcal{X}}_{t+1:t+H}\) and ground truth \(\mathcal{X}_{t+1:t+H}\), the rollout loss is
\begin{equation}
    \mathcal{L}_{\mathrm{roll}}
    =
    \frac{1}{H}
    \sum_{k=1}^{H}
    \left(
    \lambda_q
    \|\hat q_{t+k} - q_{t+k}\|_2^2
    +
    \lambda_{\dot q}
    \|\hat{\dot q}_{t+k} - \dot q_{t+k}\|_2^2
    \right),
\end{equation}
where \(\lambda_q\) and \(\lambda_{\dot q}\) balance position and velocity prediction errors.

When ground-truth velocity at time \(t\) is available, we additionally supervise the velocity head:
\begin{equation}
    \mathcal{L}_{v0}
    =
    \|\hat{\dot q}_t - \dot q_t\|_2^2 .
\end{equation}
This encourages a physically meaningful rollout initialization.

We further add energy regularization for physical consistency.
Let \(E(q,\dot q)\) denote mechanical energy:
\begin{equation}
    \mathcal{L}_{E}
    =
    \frac{1}{H}
    \sum_{k=1}^{H}
    \left\|
    E(\hat q_{t+k},\hat{\dot q}_{t+k})
    -
    E(q_{t+k},\dot q_{t+k})
    \right\|_2^2 .
\end{equation}

The final training objective is
\begin{equation}
    \mathcal{L}
    =
    \mathcal{L}_{\mathrm{roll}}
    +
    \lambda_{v0}\mathcal{L}_{v0}
    +
    \lambda_E \mathcal{L}_{E},
\end{equation}
where \(\lambda_{v0}\) and \(\lambda_E\) control initial velocity supervision and energy consistency.
All losses are computed after differentiable RK4 rollout, enabling joint optimization of the encoder, velocity head, and Lagrangian dynamics.

\subsection{Implementation Details}
\label{sec:implementation_details}

Unless otherwise specified, we use history length \(L=8\), prediction horizon \(H=32\), and time interval \(\Delta t=0.05\).
The encoder is a one-layer GRU with hidden dimension 64, followed by an MLP that outputs a 32-dimensional context.
The velocity head is a lightweight MLP taking \([q_t,z_t]\) as input.

The context-conditioned mass and potential networks are MLPs with hidden dimension 128 and Tanh activations.
Mass positivity is enforced by softplus with \(\epsilon=10^{-3}\).
Training and evaluation use RK4, and gradients are backpropagated through the full rollout without detaching intermediate states.

We optimize the model using Adam with learning rate \(10^{-3}\), batch size 256, gradient clipping of 1.0, and early stopping with patience 15 over at most 50 epochs.
The default loss weights are
\begin{equation}
    \lambda_q = 1.0, \quad
    \lambda_{\dot q} = 0.1, \quad
    \lambda_{v0} = 0.1, \quad
    \lambda_E = 0.01 .
\end{equation}
All results are evaluated under open-loop rollout using only observed history.
\section{Experiments}

\subsection{Experimental Setup}
\label{sec:experimental_setup}

\textbf{Datasets.}
We evaluate HiLNN on three pendulum-based systems: a fixed conservative pendulum, a fixed damped pendulum, and a variable-parameter pendulum with trajectory-dependent dynamics.
For all datasets, the model observes only a position history of length \(L=8\) and predicts \(H=32\) future steps with \(\Delta t=0.05\).
Details are summarized in Table~\ref{tab:dataset_setup}.

\begin{table}[!htbp]
\centering
\caption{Dataset setup with $L=8$, $H=32$, and $\Delta t=0.05$.}
\label{tab:dataset_setup}
\resizebox{\linewidth}{!}{
\begin{tabular}{llrrll}
\toprule
Dataset & Type & Train/Val/Test & Test win. & Params & Property \\
\midrule
Pendulum & Fixed conservative & 1000/200/200 & 32200 & fixed $l{=}1,m{=}1,c{=}0$ & Conservative \\
Damped & Fixed dissipative & 1000/200/200 & 32200 & fixed $c{=}0.15$ & Decay \\
Variable & Variable params. & 3000/500/500 & 80500 & sampled $l,m,c$ & Heterogeneous \\
\bottomrule
\end{tabular}
}
\end{table}

\textbf{Evaluation protocol.}
All methods follow the same position-only forecasting protocol.
At test time, each model observes only the position history and performs open-loop rollout without future ground truth.
We report MSE, MAE, Final MSE@32, and Energy MSE to measure trajectory accuracy, long-horizon error accumulation, and physical consistency.

\textbf{Baselines.}
We compare HiLNN with representative dynamics models, including LNN~\cite{cranmer2020lagrangian}, HNN~\cite{greydanus2019hamiltonian}, Neural ODE~\cite{chen2018neural}, MLP-based predictors, and an LNN-multistep variant trained with rollout supervision.
These baselines cover both black-box and physics-guided dynamics, enabling direct comparison with the proposed history-informed Lagrangian formulation.

\textbf{Implementation details.}
HiLNN uses a one-layer GRU to encode position history into a latent context, which is shared by the initial velocity head and the context-conditioned Lagrangian module.
The mass and potential terms are modeled by MLPs, with mass positivity enforced by softplus and a small constant.
Unless otherwise specified, training and evaluation use RK4 with gradients propagated through the full rollout horizon.
The training configuration is shown in Table~\ref{tab:training_config}.

\begin{table}[!htbp]
\centering
\small
\caption{Training configuration of HiLNN.}
\label{tab:training_config}
\setlength{\tabcolsep}{4pt}
\renewcommand{\arraystretch}{1.05}
\resizebox{\linewidth}{!}{
\begin{tabular}{llll}
\toprule
Item & Value & Item & Value \\
\midrule
History length $L$ & 8 
& Prediction horizon $H$ & 32 \\

Time step $\Delta t$ & 0.05 
& Encoder & GRU \\

GRU hidden dim. & 64 
& Context dim. $d_z$ & 32 \\

GRU layers & 1 
& Lagrangian MLP hidden & 128 \\

Mass positivity $\varepsilon$ & $10^{-3}$ 
& Integrator & RK4 \\

Detach between steps & False 
& Optimizer & Adam \\

Learning rate & $10^{-3}$ 
& Batch size & 256 \\

Max epochs & 50 
& Early-stop patience & 15 \\

Gradient clip norm & 1.0 
& Loss weights 
& $\lambda_q/\lambda_{\dot{q}}/\lambda_{v0}/\lambda_E=1.0/0.1/0.1/0.01$ \\
\bottomrule
\end{tabular}
}
\end{table}
\subsection{Main Quantitative Results}
\label{sec:main_results}

\begin{figure}[t]
    \centering
    \includegraphics[width=\linewidth]{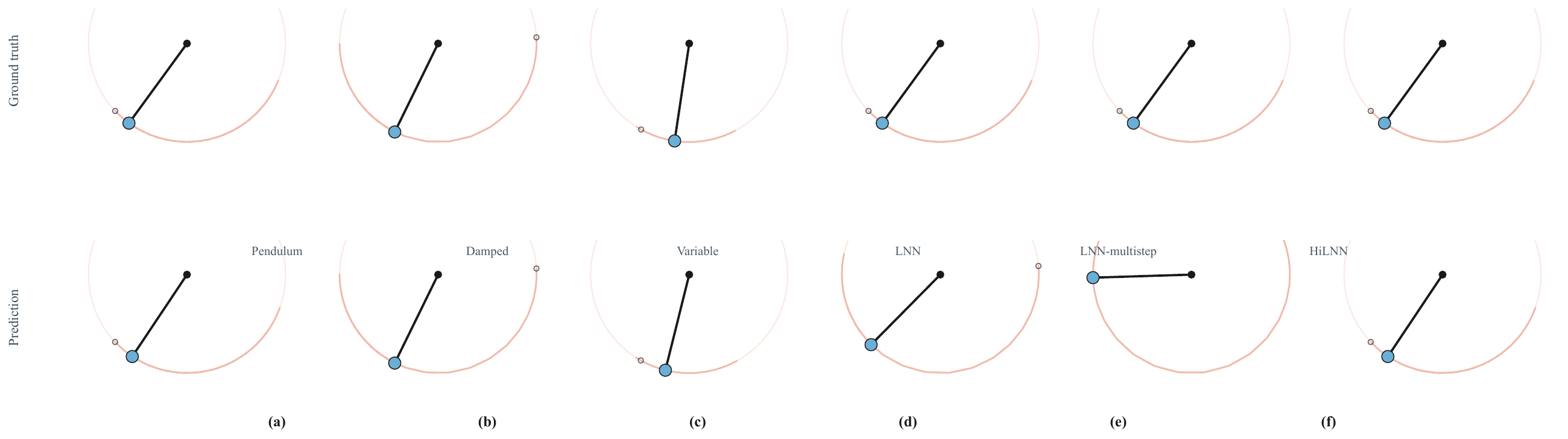}
    \caption{%
        Qualitative rollout comparison.
        Left: ground truth and HiLNN on three datasets (a--c).
        Right: ground truth and baseline predictions on the fixed pendulum (d--f).
    }
    \label{fig:schematic_qualitative}
\end{figure}

Fig.~\ref{fig:schematic_qualitative} qualitatively compares $32$-step open-loop rollouts: panels~(a--c) show ground-truth and HiLNN trajectories on the fixed, damped, and variable pendulum datasets, while panels~(d--f) contrast LNN, LNN-multistep, and HiLNN on the fixed pendulum, with HiLNN tracking the ground truth most closely. Table~\ref{tab:main_results_full} reports the overall comparison on the three pendulum systems.
HiLNN achieves the best performance across conservative, dissipative, and variable-parameter settings.
On the standard pendulum, it reduces the average MSE from 0.279 of LNN to 0.103, while also obtaining the lowest Final MSE@32 and Energy MSE, showing improved prediction accuracy and long-horizon physical consistency.

The advantage is more pronounced on damped and variable-parameter systems.
For the damped pendulum, HiLNN achieves an average MSE of \(4.28\times10^{-3}\) and reduces Energy MSE from 0.987 to 0.107 compared with LNN, indicating that the context-conditioned damping term captures dissipative dynamics.
For the variable-parameter pendulum, HiLNN reduces MSE from 0.894 of LNN-multistep to 0.061 and Final MSE@32 from 2.295 to 0.193, suggesting that the latent context effectively adapts the structured dynamics to trajectory-dependent physical variations.

Overall, these results show that a fixed global Lagrangian model is insufficient under position-only and heterogeneous dynamics.
By inferring the missing initial velocity and conditioning Lagrangian components on history, HiLNN provides more accurate and physically coherent long-horizon forecasts.

\begin{table}[!htbp]
\centering
\small
\caption{Full quantitative results on the three pendulum systems.}
\label{tab:main_results_full}
\begin{tabular}{llcccc}
\toprule
Dataset & Method & MSE $\downarrow$ & MAE $\downarrow$ & Final MSE@32 $\downarrow$ & Energy MSE $\downarrow$ \\
\midrule
\multirow{6}{*}{Pendulum} 
& LNN~\cite{cranmer2020lagrangian} & 0.279 & 0.116 & 0.483 & 4.222 \\
& LNN~\cite{cranmer2020lagrangian}-multistep & 0.415 & 0.238 & 0.857 & 5.906 \\
& HNN~\cite{greydanus2019hamiltonian} & 1.646 & 0.826 & 2.740 & 17.512 \\
& Neural~ODE~\cite{chen2018neural} & 0.580 & 0.458 & 1.355 & 9.603 \\
& MLP-one-step & 0.261 & 0.156 & 0.564 & 3.695 \\
& \ourscell{\textbf{HiLNN}} 
& \ourscell{\best{0.103}} 
& \ourscell{\best{0.109}} 
& \ourscell{\best{0.271}} 
& \ourscell{\best{1.705}} \\
\midrule
\multirow{3}{*}{Damped} 
& LNN~\cite{cranmer2020lagrangian} & 0.037 & 0.104 & 0.149 & 0.987 \\
& LNN~\cite{cranmer2020lagrangian}-multistep & 0.042 & 0.124 & 0.089 & 1.338 \\
& \ourscell{\textbf{HiLNN}} 
& \ourscell{\best{$4.28 \times 10^{-3}$}} 
& \ourscell{\best{0.036}} 
& \ourscell{\best{0.018}} 
& \ourscell{\best{0.107}} \\
\midrule
\multirow{3}{*}{Variable} 
& LNN~\cite{cranmer2020lagrangian} & 0.960 & 0.536 & 2.483 & 19.516 \\
& LNN-multistep~\cite{cranmer2020lagrangian} & 0.894 & 0.522 & 2.295 & 26.944 \\
& \ourscell{\textbf{HiLNN}} 
& \ourscell{\best{0.061}} 
& \ourscell{\best{0.113}} 
& \ourscell{\best{0.193}} 
& \ourscell{\best{1.759}} \\
\bottomrule
\end{tabular}
\end{table}
\subsection{Long-Horizon Rollout Analysis}
\label{sec:rollout_analysis}

\begin{figure}[t]
    \centering
    \includegraphics[width=0.98\textwidth]{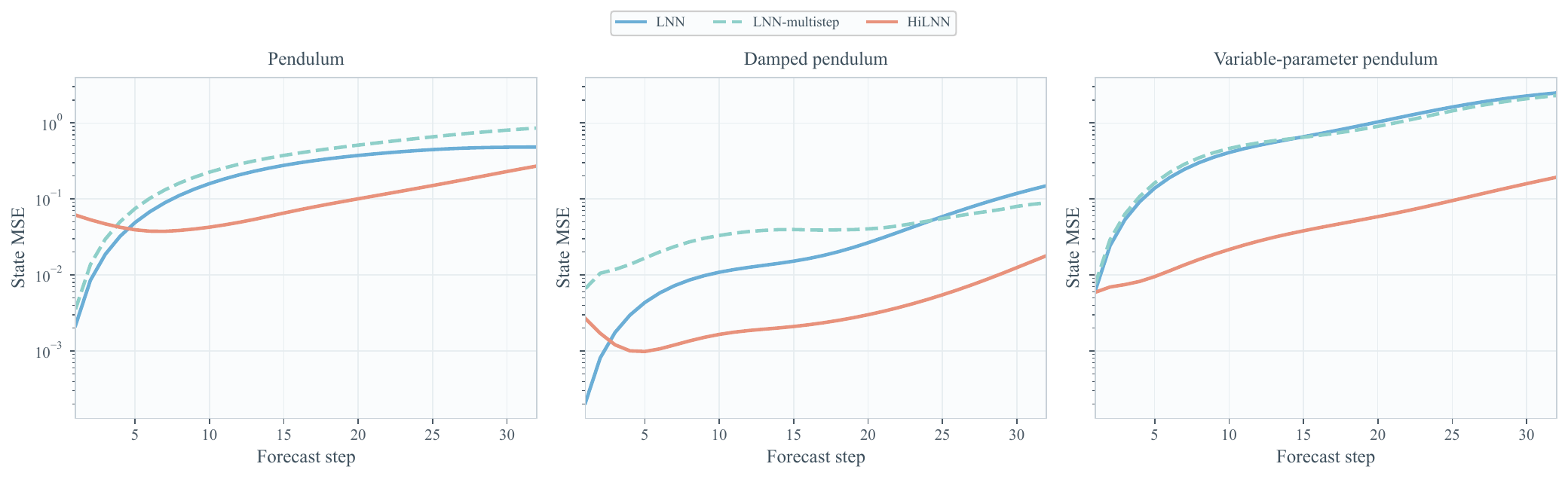}
    \caption{
    Step-wise rollout error on the three pendulum systems.
    HiLNN achieves lower state MSE than LNN and LNN-multistep over the 32-step open-loop prediction horizon.
    }
    \label{fig:rollout_mse_three_systems}
\end{figure}

To further analyze error accumulation, we report step-wise rollout errors over the 32-step open-loop horizon.
As shown in Fig.~\ref{fig:rollout_mse_three_systems}, HiLNN consistently achieves lower errors on all three systems, especially at medium and long horizons.
This shows that the history-informed context improves prediction accuracy and mitigates recursive error growth in open-loop forecasting.

\begin{figure*}[t]
    \centering
    \includegraphics[width=0.98\textwidth]{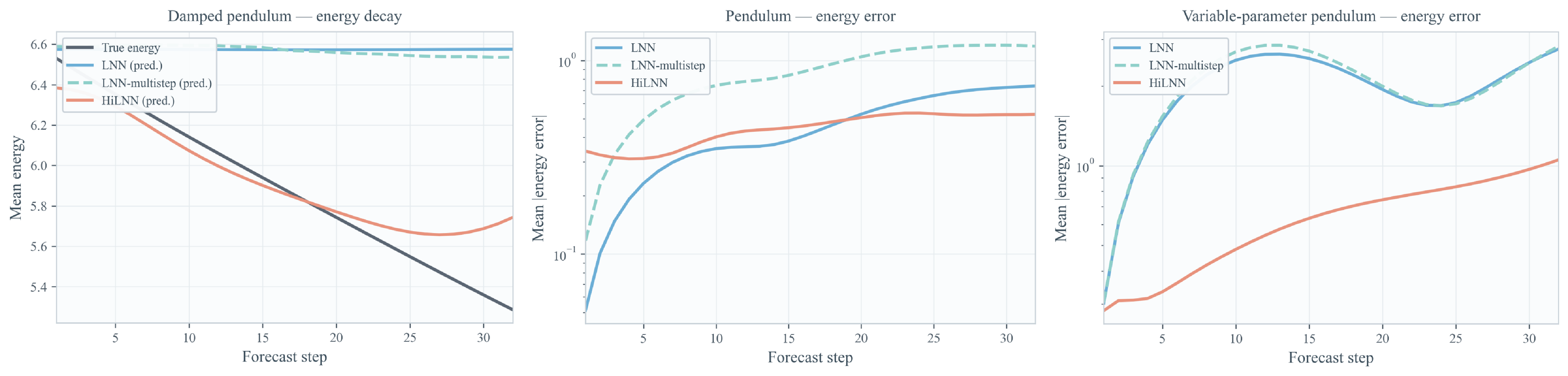}
    \caption{
    Energy behavior on the damped, standard, and variable-parameter pendulum systems.
    }
    \label{fig:energy_behavior_three_systems}
\end{figure*}

Table~\ref{tab:rollout_summary_pendulum} reports detailed rollout errors on the conservative pendulum.
Although LNN has the lowest one-step error, its error grows quickly over longer horizons.
HiLNN achieves lower errors from Step 8 onward and reduces the final-step error from 0.483 to 0.271, showing that velocity inference and history-conditioned Lagrangian dynamics improve long-horizon stability.

\begin{table}[!htbp]
\centering
\small
\caption{Rollout error summary on the pendulum system. Values report state MSE at selected forecast steps.}
\label{tab:rollout_summary_pendulum}
\begin{tabular}{lccccc}
\toprule
Method & Step 1 & Step 8 & Step 16 & Step 32 & Avg. \\
\midrule
LNN~\cite{cranmer2020lagrangian} & \best{$2.17 \times 10^{-3}$} & 0.112 & 0.299 & 0.483 & 0.279 \\
LNN~\cite{cranmer2020lagrangian}-multistep & $3.55 \times 10^{-3}$ & 0.162 & 0.402 & 0.857 & 0.415 \\
HNN~\cite{greydanus2019hamiltonian} & 0.024 & 0.897 & 1.848 & 2.740 & 1.646 \\
Neural~ODE~\cite{chen2018neural} & $8.51 \times 10^{-3}$ & 0.252 & 0.504 & 1.355 & 0.580 \\
\ourscell{\textbf{HiLNN}} 
& \ourscell{0.061} 
& \ourscell{\best{0.039}} 
& \ourscell{\best{0.072}} 
& \ourscell{\best{0.271}} 
& \ourscell{\best{0.103}} \\
\bottomrule
\end{tabular}
\end{table}

The damped pendulum results in Table~\ref{tab:rollout_summary_damped} further show HiLNN's advantage under dissipative dynamics.
Although LNN has a very small Step-1 error, its error grows rapidly at later steps.
HiLNN achieves the lowest errors at Steps 8, 16, and 32, reducing the final-step error from 0.149 to 0.018.
This indicates that context-conditioned damping better captures long-term energy decay.

\begin{table}[!htbp]
\centering
\small
\caption{Rollout error summary on the damped pendulum system. Values report state MSE at selected forecast steps.}
\label{tab:rollout_summary_damped}
\begin{tabular}{lccccc}
\toprule
Method & Step 1 & Step 8 & Step 16 & Step 32 & Avg. \\
\midrule
LNN~\cite{cranmer2020lagrangian} & \best{$2.08 \times 10^{-4}$} & $8.60 \times 10^{-3}$ & 0.016 & 0.149 & 0.037 \\
LNN~\cite{cranmer2020lagrangian}-multistep & $6.62 \times 10^{-3}$ & 0.027 & 0.039 & 0.089 & 0.042 \\
\ourscell{\textbf{HiLNN}} 
& \ourscell{$2.68 \times 10^{-3}$} 
& \ourscell{\best{$1.36 \times 10^{-3}$}} 
& \ourscell{\best{$2.21 \times 10^{-3}$}} 
& \ourscell{\best{0.018}} 
& \ourscell{\best{$4.28 \times 10^{-3}$}} \\
\bottomrule
\end{tabular}
\end{table}

The improvement is most evident in the variable-parameter setting, as shown in Table~\ref{tab:rollout_summary_variable}.
Since trajectories follow different physical parameters, a single global Lagrangian model struggles with long-horizon prediction.
HiLNN achieves the best performance at all selected steps and reduces the Step-32 error from 2.483/2.295 to 0.193 compared with LNN/LNN-multistep.
This indicates that the latent context captures trajectory-dependent dynamics and enables adaptation across heterogeneous systems.

\begin{table}[!htbp]
\centering
\small
\caption{Rollout error summary on the variable-parameter pendulum system. Values report state MSE at selected forecast steps.}
\label{tab:rollout_summary_variable}
\begin{tabular}{lccccc}
\toprule
Method & Step 1 & Step 8 & Step 16 & Step 32 & Avg. \\
\midrule
LNN~\cite{cranmer2020lagrangian} & $6.12 \times 10^{-3}$ & 0.301 & 0.717 & 2.483 & 0.960 \\
LNN~\cite{cranmer2020lagrangian}-multistep & $7.56 \times 10^{-3}$ & 0.350 & 0.685 & 2.295 & 0.894 \\
\ourscell{\textbf{HiLNN}} 
& \ourscell{\best{$5.92 \times 10^{-3}$}} 
& \ourscell{\best{0.016}} 
& \ourscell{\best{0.042}} 
& \ourscell{\best{0.193}} 
& \ourscell{\best{0.061}} \\
\bottomrule
\end{tabular}
\end{table}

Overall, these analyses show that baseline models may achieve competitive short-term prediction but suffer from rapid error accumulation in long open-loop rollouts.
HiLNN stabilizes error growth by using history to infer both the missing initial state and trajectory-specific dynamical context.
\subsection{Trajectory and Physical Consistency}
\label{sec:trajectory_physical_consistency}

\begin{figure}[t]
    \centering
    \includegraphics[width=\linewidth]{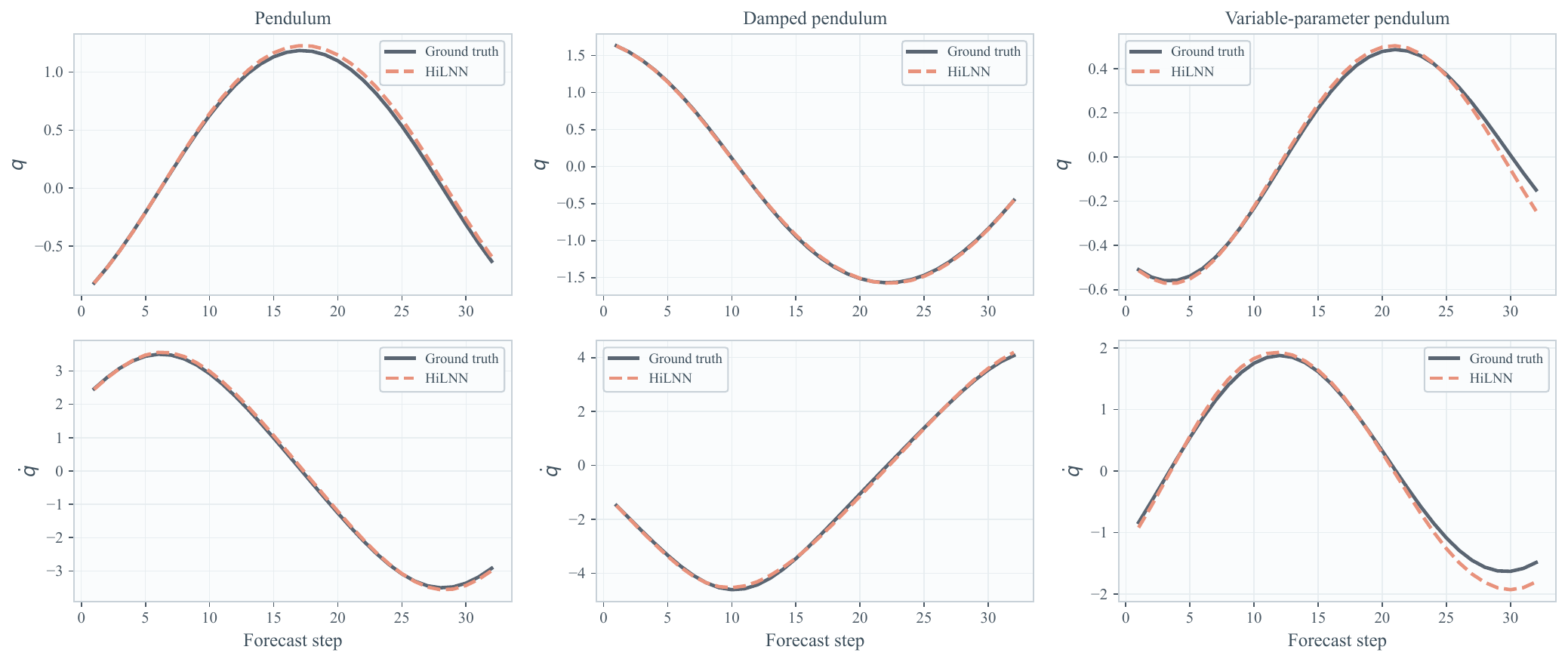}
    \caption{
    Representative trajectory predictions on the three pendulum systems.
    HiLNN closely follows the ground-truth position \(q\) and velocity \(\dot q\) over the 32-step prediction horizon.
    }
    \label{fig:trajectory_three_systems}
\end{figure}

Beyond aggregate metrics, we further examine trajectory and energy behavior.
Fig.~\ref{fig:trajectory_three_systems} shows that HiLNN closely tracks the ground-truth position \(q\) and velocity \(\dot q\) over 32 steps, whereas baselines show larger phase shifts or amplitude errors, consistent with Sec.~\ref{sec:rollout_analysis}.
For physical consistency, Table~\ref{tab:energy_summary} shows that HiLNN obtains the lowest energy errors at Steps 16 and 32 on the damped pendulum, reducing Test Energy MSE from 0.987 to 0.107 and mean absolute energy error from 0.686 to 0.167.
These results suggest that history-informed context improves initialization and rollout dynamics, while context-conditioned damping better captures dissipative energy decay.
These results indicate that history-informed context improves state initialization and dynamical evolution, while context-conditioned damping helps capture dissipative energy decay.






The variable-parameter setting further evaluates physical consistency under trajectory-dependent parameters.
As shown in Table~\ref{tab:energy_summary}, LNN and LNN-multistep accumulate large energy errors, with Test Energy MSE values of 19.516 and 26.944.
In contrast, HiLNN reduces Test Energy MSE to 1.759 and Step-32 energy error from 26.111 to 3.646 compared with LNN.
Although not best at Step 1, HiLNN achieves much lower later-step errors, showing that the learned context improves long-term physical consistency under heterogeneous dynamics.

\begin{table}[!htbp]
\centering
\small
\caption{Energy behavior summary on damped and variable-parameter pendulum systems. For the damped system, the true energy drop is approximately 1.24 over 32 steps on average.}
\label{tab:energy_summary}
\setlength{\tabcolsep}{4pt}
\renewcommand{\arraystretch}{1.08}
\resizebox{\linewidth}{!}{
\begin{tabular}{llccccc}
\toprule
Dataset & Method & $E$-MSE@1 & $E$-MSE@16 & $E$-MSE@32 & Mean $|$err$|$ & Test $E$-MSE \\
\midrule

\multirow{3}{*}{Damped}
& LNN~\cite{cranmer2020lagrangian} 
& \best{$5.31 \times 10^{-3}$} & 0.800 & 2.607 & 0.686 & 0.987 \\
& LNN~\cite{cranmer2020lagrangian}-multistep 
& 0.225 & 1.277 & 2.825 & 0.730 & 1.338 \\
& \ourscell{\textbf{HiLNN}} 
& \ourscell{0.085} 
& \ourscell{\best{0.044}} 
& \ourscell{\best{0.565}} 
& \ourscell{\best{0.167}} 
& \ourscell{\best{0.107}} \\

\midrule

\multirow{3}{*}{Variable}
& LNN~\cite{cranmer2020lagrangian} 
& \best{0.386} & 31.072 & 26.111 & 2.004 & 19.516 \\
& LNN~\cite{cranmer2020lagrangian}-multistep 
& 0.462 & 44.162 & 31.178 & 2.073 & 26.944 \\
& \ourscell{\textbf{HiLNN}} 
& \ourscell{0.530} 
& \ourscell{\best{1.675}} 
& \ourscell{\best{3.646}} 
& \ourscell{\best{0.645}} 
& \ourscell{\best{1.759}} \\

\bottomrule
\end{tabular}
}
\end{table}

Overall, these results show that HiLNN improves both trajectory accuracy and physical consistency.
By following dissipative energy decay and reducing long-horizon energy drift in parameter-varying systems, HiLNN validates the benefit of combining history-informed context with structured Lagrangian dynamics.

\subsection{Ablation Studies}
\label{sec:ablation_studies}

We conduct ablations on energy regularization, rollout training, initial velocity supervision, and history length.


\begin{table}[!htbp]
\centering
\small
\caption{Ablation summary of HiLNN on the pendulum system, covering energy regularization, rollout training design, and initial velocity supervision. ``--'' indicates that the metric is not applicable to the corresponding ablation.}
\label{tab:ablation_summary}
\setlength{\tabcolsep}{4pt}
\renewcommand{\arraystretch}{1.08}
\resizebox{\linewidth}{!}{
\begin{tabular}{llcccc}
\toprule
Ablation & Setting & MSE $\downarrow$ & Final MSE@32 $\downarrow$ & Energy MSE $\downarrow$ & Init. Vel. MSE $\downarrow$ \\
\midrule

\multirow{4}{*}{Energy weight}
& $\lambda_E=0$ & 0.118 & 0.261 & 3.147 & -- \\
& $\lambda_E=0.001$ & 0.116 & \best{0.253} & 2.680 & -- \\
& \ourscell{\textbf{$\lambda_E=0.01$}} 
& \ourscell{\best{0.103}} 
& \ourscell{0.271} 
& \ourscell{\best{1.705}} 
& \ourscell{--} \\
& $\lambda_E=0.05$ & 0.261 & 0.495 & 4.041 & -- \\

\midrule
\multirow{3}{*}{Rollout design}
& HiLNN v1a, Euler, detach & 0.449 & 1.025 & 9.070 & -- \\
& \ourscell{\textbf{HiLNN v1b, RK4, full BPTT}} 
& \ourscell{\best{0.118}} 
& \ourscell{\best{0.261}} 
& \ourscell{\best{3.147}} 
& \ourscell{--} \\
& LNN~\cite{cranmer2020lagrangian}, RK4 & 0.279 & 0.482 & 4.222 & -- \\

\midrule
\multirow{2}{*}{Init. velocity}
& \ourscell{\textbf{$\lambda_{v0}=0.1$}} 
& \ourscell{0.118} 
& \ourscell{\best{0.261}} 
& \ourscell{\best{3.147}} 
& \ourscell{\best{0.115}} \\
& $\lambda_{v0}=0$ & \best{0.117} & 0.262 & 3.318 & 0.245 \\

\bottomrule
\end{tabular}
}
\end{table}

Unless otherwise specified, all ablations are performed on the pendulum setting.
Table~\ref{tab:ablation_summary} summarizes the effects of energy regularization, rollout design, and initial velocity supervision.
For energy regularization, removing \(\lambda_E\) weakens physical consistency, yielding an MSE of 0.118 and an Energy MSE of 3.147, while a small weight improves the accuracy--energy trade-off.
Specifically, \(\lambda_E=0.01\) gives the best overall MSE and Energy MSE, whereas \(\lambda_E=0.001\) achieves the lowest Final MSE@32 but higher Energy MSE.
A larger weight, \(\lambda_E=0.05\), hurts both accuracy and energy consistency, indicating over-regularization; thus, we use \(\lambda_E=0.01\) by default.
For rollout training, the Euler variant with detached states, HiLNN v1a, performs poorly, with an MSE of 0.449 and Final MSE@32 of 1.025.
Using RK4 with full backpropagation through time reduces them to 0.118 and 0.261, respectively, and also outperforms the LNN baseline under the same RK4 protocol.
This confirms the benefit of differentiable high-order integration and full-horizon gradient propagation.
Finally, removing the auxiliary initial velocity loss keeps rollout MSE similar but increases the initial velocity error from 0.115 to 0.245, showing that \(\mathcal{L}_{v0}\) mainly improves the accuracy and interpretability of the inferred initial state.
We retain this loss to provide a more physical initialization for open-loop rollout.

\section{Conclusion}

In this paper, we proposed HiLNN, a history-informed Lagrangian framework for long-horizon mechanical forecasting from position-only observations. By inferring a latent context from observed history, HiLNN estimates the missing initial velocity and adaptively conditions the mass, potential, and damping terms of a structured Lagrangian model. Combined with differentiable RK4 rollout and multi-step supervision, HiLNN preserves mechanical structure while improving long-horizon stability. Experiments on conservative, dissipative, and variable-parameter pendulum systems show that HiLNN achieves more accurate and physically consistent predictions than representative black-box and physics-guided baselines. 

\section*{Acknowledgements}

The work is supported by the National Natural Science Foundation of
China under Grant No.~62076078, the Fundamental Research Funds for
the Central Universities under Grant No.~3072024LJ0403, and the
CAAI-Huawei MindSpore Open Fund under Grant
No.~CAAI--XSJLJJ--2020--033A.


\bibliographystyle{splncs04}
\bibliography{references}

@inproceedings{chen2018neural,
  title={Neural Ordinary Differential Equations},
  author={Chen, Ricky T. Q. and Rubanova, Yulia and Bettencourt, Jesse and Duvenaud, David K.},
  booktitle={Advances in Neural Information Processing Systems},
  volume={31},
  pages={6572--6583},
  year={2018}
}

@inproceedings{rubanova2019latent,
  title={Latent ODEs for Irregularly-Sampled Time Series},
  author={Rubanova, Yulia and Chen, Ricky T. Q. and Duvenaud, David},
  booktitle={Advances in Neural Information Processing Systems},
  volume={32},
  year={2019}
}

@inproceedings{greydanus2019hamiltonian,
  title={Hamiltonian Neural Networks},
  author={Greydanus, Samuel and Dzamba, Misko and Yosinski, Jason},
  booktitle={Advances in Neural Information Processing Systems},
  volume={32},
  pages={15353--15363},
  year={2019}
}

@inproceedings{cranmer2020lagrangian,
  title={Lagrangian Neural Networks},
  author={Cranmer, Miles and Greydanus, Sam and Hoyer, Stephan and Battaglia, Peter and Spergel, David and Ho, Shirley},
  booktitle={ICLR 2020 Workshop on Integration of Deep Neural Models and Differential Equations},
  year={2020}
}

@inproceedings{lutter2019deep,
  title={Deep Lagrangian Networks: Using Physics as Model Prior for Deep Learning},
  author={Lutter, Michael and Ritter, Christian and Peters, Jan},
  booktitle={International Conference on Learning Representations},
  year={2019}
}

@article{krishnan2015deep,
  title={Deep Kalman Filters},
  author={Krishnan, Rahul G. and Shalit, Uri and Sontag, David},
  journal={arXiv preprint arXiv:1511.05121},
  year={2015}
}

@inproceedings{yildiz2019ode2vae,
  title={ODE2VAE: Deep Generative Second Order ODEs with Bayesian Neural Networks},
  author={Yildiz, Cagatay and Heinonen, Markus and Lahdesmaki, Harri},
  booktitle={Advances in Neural Information Processing Systems},
  volume={32},
  year={2019}
}

@article{zhang2025floating,
  title={Floating-Body Hydrodynamic Neural Networks},
  author={Zhang, Tianshuo and Zhai, Wenzhe and Yann, Rui and Gao, Jia and Cao, He and Xing, Xianglei},
  journal={arXiv preprint arXiv:2509.13783},
  year={2025}
}

@inproceedings{battaglia2016interaction,
  title={Interaction Networks for Learning about Objects, Relations and Physics},
  author={Battaglia, Peter W. and Pascanu, Razvan and Lai, Matthew and Rezende, Danilo Jimenez and Kavukcuoglu, Koray},
  booktitle={Advances in Neural Information Processing Systems},
  volume={29},
  pages={4502--4510},
  year={2016}
}

@inproceedings{nagabandi2018neural,
  title={Neural Network Dynamics for Model-Based Deep Reinforcement Learning with Model-Free Fine-Tuning},
  author={Nagabandi, Anusha and Kahn, Gregory and Fearing, Ronald S. and Levine, Sergey},
  booktitle={2018 IEEE International Conference on Robotics and Automation},
  pages={7559--7566},
  year={2018},
  doi={10.1109/ICRA.2018.8463189}
}

@inproceedings{sanchez2020learning,
  title={Learning to Simulate Complex Physics with Graph Networks},
  author={Sanchez-Gonzalez, Alvaro and Godwin, Jonathan and Pfaff, Tobias and Ying, Rex and Leskovec, Jure and Battaglia, Peter W.},
  booktitle={Proceedings of the 37th International Conference on Machine Learning},
  volume={119},
  pages={8459--8468},
  year={2020}
}

@inproceedings{finzi2020simplifying,
  title={Simplifying Hamiltonian and Lagrangian Neural Networks via Explicit Constraints},
  author={Finzi, Marc and Wang, Ke Alexander and Wilson, Andrew Gordon},
  booktitle={Advances in Neural Information Processing Systems},
  volume={33},
  pages={13880--13889},
  year={2020}
}

@inproceedings{zhong2020symplectic,
  title={Symplectic ODE-Net: Learning Hamiltonian Dynamics with Control},
  author={Zhong, Yaofeng Desmond and Dey, Biswadip and Chakraborty, Amit},
  booktitle={International Conference on Learning Representations},
  year={2020}
}

@article{sosanya2022dissipative,
  title={Dissipative Hamiltonian Neural Networks: Learning Dissipative and Conservative Dynamics Separately},
  author={Sosanya, Andrew and Greydanus, Sam},
  journal={arXiv preprint arXiv:2201.10085},
  year={2022}
}

@article{desai2021port,
  title={Port-Hamiltonian Neural Networks for Learning Explicit Time-Dependent Dynamical Systems},
  author={Desai, Shaan A. and Mattheakis, Marios and Sondak, David and Protopapas, Pavlos and Roberts, Stephen J.},
  journal={Physical Review E},
  volume={104},
  number={3},
  pages={034312},
  year={2021},
  doi={10.1103/PhysRevE.104.034312}
}

@incollection{takens1981detecting,
  title={Detecting Strange Attractors in Turbulence},
  author={Takens, Floris},
  booktitle={Dynamical Systems and Turbulence, Warwick 1980},
  series={Lecture Notes in Mathematics},
  volume={898},
  pages={366--381},
  publisher={Springer},
  address={Berlin, Heidelberg},
  year={1981},
  doi={10.1007/BFb0091924}
}

@article{buissonfenet2023recognition,
  title={Recognition Models to Learn Dynamics from Partial Observations with Neural ODEs},
  author={Buisson-Fenet, Mona and Morgenthaler, Valery and Trimpe, Sebastian and Di Meglio, Florent},
  journal={Transactions on Machine Learning Research},
  year={2023}
}

@article{li2025trajectory,
  title={Trajectory Prediction Methods Based on Analytical Mechanics and Graph Neural Networks},
  author={Li, Minghan and Xiao, Yang and Xing, Xianglei},
  journal={CAAI Transactions on Intelligent Systems},
  volume={20},
  number={6},
  pages={1355--1365},
  year={2025},
  doi={10.11992/tis.202501020}
}

@inproceedings{li2025frequency,
  title={Frequency-Aligned Knowledge Distillation for Lightweight Spatiotemporal Forecasting},
  author={Li, Yuqi and Yang, Chuanguang and Zeng, Hansheng and Dong, Zeyu and An, Zhulin and Xu, Yongjun and Tian, Yingli and Wu, Hao},
  booktitle={Proceedings of the IEEE/CVF International Conference on Computer Vision},
  pages={7262--7272},
  year={2025}
}

@article{li2026evolving,
  title={Evolving Multimodal Models for Physical Dynamics: A Multi-Objective Neuroevolution Approach},
  author={Li, Yuqi and Dong, Junhao and Liu, Jiao and Koniusz, Piotr and Zeng, Hansheng and Yang, Chuanguang and Liu, Junming and Tian, Yingli and Huang, Tingwen and Wu, Hao},
  journal={IEEE Transactions on Evolutionary Computation},
  year={2026},
  doi={10.1109/TEVC.2026.3698641}
}

@inproceedings{li2026distilling,
  title={Distilling Time Series Foundation Models for Efficient Forecasting},
  author={Li, Yuqi and Ding, Kuiye and Yang, Chuanguang and Chen, Szu-Yu and Tian, Yingli},
  booktitle={IEEE International Conference on Acoustics, Speech and Signal Processing},
  year={2026}
}

@article{li2025ddtime,
  title={DDTime: Dataset Distillation with Spectral Alignment and Information Bottleneck for Time-Series Forecasting},
  author={Li, Yuqi and Ding, Kuiye and Yang, Chuanguang and Wang, Hao and Wang, Haoxuan and Duan, Huiran and Liu, Junming and Tian, Yingli},
  journal={arXiv preprint arXiv:2511.16715},
  year={2025}
}
\end{document}